\documentclass[runningheads]{llncs}

\usepackage{eccv}

\usepackage{eccvabbrv}

\usepackage{graphicx}
\usepackage{orcidlink}
\usepackage{booktabs}
\usepackage{multirow}
\usepackage{amsmath,amssymb}
\usepackage{microtype}
\usepackage{hyperref}
\usepackage{orcidlink}
\usepackage{xcolor}
\usepackage{enumitem}
\usepackage{tabularx}
\usepackage{adjustbox}
\usepackage[accsupp]{axessibility}

\newif\ifdraft
\draftfalse

\title{Listen Before You Speak: Response Planning from Listener Facial Reactions for Conversational Speech Generation}
\titlerunning{Response Planning from Listener Facial Reactions}

\author{Yunji Chu\orcidlink{0009-0003-8348-1312}}
\authorrunning{Y.~Chu}
\institute{Department of Artificial Intelligence, Sogang University, Seoul, Republic of Korea\\
\email{yungie222@sogang.ac.kr}}

\begin{document}
\maketitle

\begin{abstract}
Conversational speech depends on dialogue context and the listener's immediately preceding behavior. We propose ReACT-TTS, a two-stage framework that uses a one-second pre-response listener facial sequence to plan the next utterance's emotion and prosody before speech realization. On a strict dyadic MELD protocol, Temporal conditioning yields higher mean macro-F1 and VAD concordance than Text-only across ten seeds, while accuracy remains essentially unchanged. Ablations show that temporal modeling performs best among the visual variants and that an explicit early-to-late difference is unnecessary; correct listener reactions also outperform cyclic mismatches on average. In a contextual-appropriateness study with 20 speech researchers, 76\% of judgments prefer Temporal, 9\% Text-only, and 15\% report no preference. We further connect the predicted response style to a Grad-TTS backbone for end-to-end speech realization. Overall, the results support pre-response listener dynamics as complementary cues for conversational response planning. The source code is available at \href{https://github.com/CYJ1/ReACT-TTS_public}{ReACT-TTS}.
\keywords{Conversational speech generation \and Listener facial reaction \and Response planning \and Audio-visual learning \and Expressive TTS}
\end{abstract}

\section{Introduction}

A spoken response is shaped by both lexical content and interaction. Speakers monitor others and adjust vocal delivery accordingly; visible amusement, discomfort, or confusion can alter emotion, pitch, energy, timing, and speaking rate even when the next sentence is fixed. Conversational speech generation should therefore model how the immediately preceding interaction informs delivery.

Conversational TTS conditions the current utterance on previous turns, speaker roles, acoustic context, or inferred affect. Dialogue history is useful, but it does not fully capture non-verbal listener behavior: a verbally neutral listener may still show surprise, skepticism, amusement, discomfort, or disengagement. Such cues can provide complementary evidence for selecting the delivery of the next response.

Visual conditioning has also been explored in speech synthesis, primarily through images of the target speaker. In contrast, our visual input depicts the \emph{listener}. Importantly, the listener reaction is not treated as an emotion label that should be copied to the target speaker. The appropriate response depends jointly on the target text and conversational context: the same visible reaction may invite an apology, reassurance, explanation, refusal, or playful continuation.

Visual-Aware TTS (VA-TTS)~\cite{zhou2023visualaware} previously demonstrated that sequential listener feedback can condition speech synthesis. ReACT-TTS focuses on a complementary question: whether the listener's facial behavior immediately before the target response can improve an explicit response-planning stage before acoustic generation. This factorization lets us measure the visual contribution independently of synthesis quality and then test whether the planned style can be connected to an end-to-end TTS system.

We formulate listener-aware response planning using the previous three dialogue turns, the target response text, and a one-second listener face sequence immediately preceding the target response. The listener sequence is encoded temporally and fused with linguistic context through target-conditioned gating. Rather than making an explicit frame-difference feature central to the method, our final model relies on temporal encoding itself; a controlled ablation evaluates whether explicit difference features add useful information.

The revised experimental protocol addresses three questions. First, does temporal listener conditioning provide complementary information over text context alone? Second, does the model depend on the \emph{correct} listener reaction rather than generic visual input? Third, can the resulting response representation be connected to speech generation without changing the underlying text or target speaker? We answer these questions using a strictly filtered MELD protocol with fixed train/dev/test criteria, multi-seed evaluation, a correct-versus-mismatched listener intervention, and end-to-end Grad-TTS generation.

The main contributions are:
\begin{itemize}[leftmargin=*,nosep]
    \item A response-planning formulation that uses the listener's one-second \emph{pre-response temporal facial reaction} as complementary context for conversational speech generation, rather than directly mirroring listener emotion.
    \item A strict dyadic MELD protocol yielding 1,117/116/261 train/dev/test response-planning samples, with all filtering thresholds determined without test-set tuning.
    \item Multi-seed controlled experiments showing a positive tendency from temporal listener conditioning, including static-versus-temporal and explicit-difference ablations, plus a mismatched-listener intervention that probes listener-specific information use.
    \item A perceptual contextual-appropriateness study with 20 speech researchers, together with end-to-end Grad-TTS evaluation on 252 test utterances, without claiming superior acoustic quality.
\end{itemize}

Code is publicly available on GitHub at \href{https://github.com/CYJ1/ReACT-TTS_public}{ReACT-TTS\_public}.

\section{Related Work}

\subsection{Face-Conditioned and Visually Guided Speech Synthesis}

Visual conditioning has been explored in TTS primarily to infer the voice or expressive characteristics of the target speaker. Face-TTS learns a face-styled diffusion model with a cross-modal biometric objective~\cite{lee2023imaginary}; Face-StyleSpeech separates face-derived speaker information from residual prosody~\cite{kang2025facestylespeech}; and FEIM-TTS combines facial representation with explicit emotion-intensity control~\cite{chu2024feim}. More recent systems broaden visual conditioning to diverse portrait styles or full-face expressive cues. FaceSpeak suppresses visually irrelevant appearance factors while extracting identity and emotion information~\cite{zhang2025facespeak}, whereas AVLM integrates full-face visual representations into an expressive speech language model for emotion-aware generation~\cite{tan2025seeing}. These methods primarily use the depicted face to characterize the target speaker or the observed speaker. ReACT-TTS instead uses the listener's reaction to plan the target speaker's next delivery.

The closest prior work is Visual-Aware TTS (VA-TTS)~\cite{zhou2023visualaware}, which conditions speech synthesis on sequential visual feedback from a listener. ReACT-TTS builds on this formulation but replaces direct visual--acoustic fusion with an explicit response-planning stage that combines dialogue semantics with temporal facial dynamics. A mismatched-listener counterfactual intervention further tests whether response planning actually depends on the observed listener.

\subsection{Conversational Speech Synthesis}

Conversational speech synthesis generates a target utterance using preceding turns rather than treating each sentence independently. DailyTalk introduced a spoken-dialogue corpus and contextual TTS baseline~\cite{lee2023dailytalk}. Later systems model intra- and inter-modal context interaction~\cite{jia2024i3css}, fine-grained semantic and prosodic graphs~\cite{jia2025mfcig}, and diffusion-based context-aware prosody~\cite{wu2025diffcss}. Recent work also makes emotion inference more explicit: prompt-guided emotive TTS derives contextual emotion tags and localized acoustic cues~\cite{jeon2025prompt}, while Chain-Talker separates emotion understanding, semantic understanding, and empathetic rendering~\cite{hu2025chaintalker}. UniTalker extends contextual synthesis toward joint speech and talking-face generation~\cite{hu2025unitalker}. These studies establish the value of dialogue history, but they do not explicitly evaluate the immediately pre-response listener reaction as a separate signal for planning the target speaker's delivery.

\subsection{Listener-Reaction Modeling in Dyadic Interaction}

A complementary line of work generates listener behavior from a speaker's verbal and non-verbal signals. ReactFace models multiple appropriate and synchronized facial reactions instead of a single deterministic response~\cite{luo2024reactface}. The REACT 2025 challenge further formalizes the one-to-many nature of listener reactions and introduces the large-scale MARS benchmark for dyadic interaction~\cite{song2025react}. ReactDiff combines multimodal interaction modeling with latent diffusion to generate diverse, contextually appropriate facial reactions~\cite{li2025reactdiff}. Although these methods generate the listener rather than the next speaker's voice, they support treating listener behavior as a dynamic, one-to-many contextual signal rather than a label that should be directly mirrored.

\subsection{Multimodal Emotion in Conversation}

MELD contains approximately 13,000 utterances from 1,433 multi-party dialogues with aligned text, audio, video, speaker, emotion, and sentiment annotations~\cite{poria2019meld}. We use it as a source of audiovisual conversational sequences rather than as a standard utterance-level emotion-recognition benchmark. Because MELD does not explicitly annotate the listener or addressee for every turn, the proposed subset retains only clips with a reliable non-speaking face track and unambiguous local interaction structure.

\section{Method}

\subsection{Task Definition}

At dialogue turn $t$, the target speaker produces utterance $x_t$ after the previous three dialogue turns
\begin{equation}
\mathcal{H}_t=\{(x_i,s_i)\}_{i=t-3}^{t-1},
\label{eq:history}
\end{equation}
where $x_i$ and $s_i$ denote transcript and speaker identity. In addition to the target text $x_t$, the response planner observes a listener-face sequence
\begin{equation}
V_t^L=\{I_n\}_{n=1}^{N_t}, \qquad 8\le N_t\le16,
\end{equation}
formed from the valid face tracks among 16 requested frames in the one-second interval immediately preceding the target response. This timing is central to our formulation: the visual input is evidence available \emph{before} the target speaker begins speaking, rather than a visual trace of the target utterance itself.

The goal is to first infer an appropriate response style and then realize the target speech:
\begin{align}
z_t^{\mathrm{style}} &= P(x_t,\mathcal{H}_t,V_t^L), \\
\hat a_t &= S(x_t,z_{s_t},z_t^{\mathrm{style}}),
\end{align}
where $z_{s_t}$ denotes the target-speaker representation. The response target describes the target utterance, not the listener's emotion. Consequently, listener behavior is used as contextual evidence rather than as a label to be mirrored.

\begin{figure*}[t]
    \centering
    \includegraphics[width=0.98\textwidth]{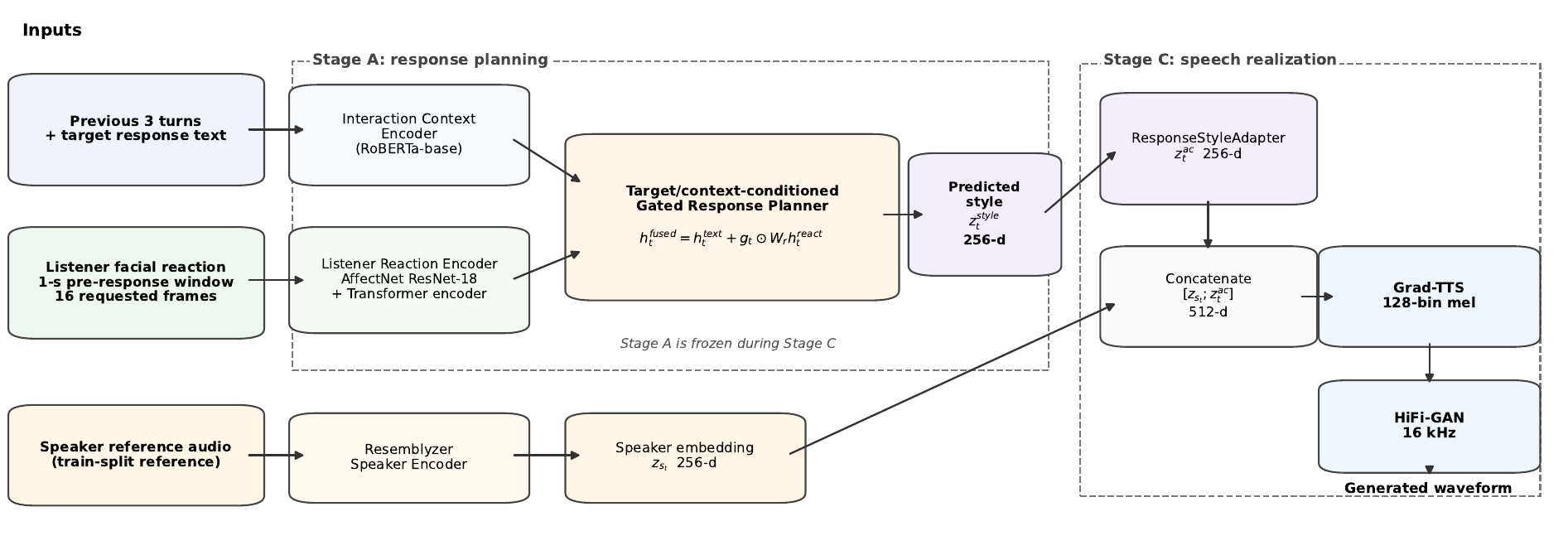}
    \caption{Overall architecture of ReACT-TTS. Dialogue context and target text are combined with the listener facial reaction from the 1-s pre-response window to predict a 256-dimensional response-style representation. During speech realization, a train-split speaker-reference waveform is encoded with Resemblyzer~\cite{wan2018ge2e}, while the predicted style is mapped by the ResponseStyleAdapter; the two 256-dimensional vectors are concatenated to condition Grad-TTS, and HiFi-GAN converts the generated 128-bin mel-spectrogram to waveform speech.}
    \label{fig:overview}
\end{figure*}

\subsection{Interaction Context Encoder}

The dialogue history is serialized with explicit speaker-role tokens:
\begin{equation}
q_t=[\mathrm{SPK}_{t-3}]x_{t-3}\cdots[\mathrm{SPK}_{t-1}]x_{t-1}[\mathrm{TARGET}]x_t.
\end{equation}
A pretrained language encoder produces token representations that are pooled into $h_t^{\mathrm{text}}$. Including the target text is important because the same listener reaction can imply different appropriate deliveries depending on whether the upcoming response is, for example, an apology, explanation, reassurance, or playful continuation.

\subsection{Temporal Listener Reaction Encoder}

Each valid frame is encoded by an expression-oriented facial backbone,
\begin{equation}
f_n=E_{\mathrm{face}}(I_n), \qquad n=1,\ldots,N_t,
\end{equation}
implemented with a ResNet-18 pretrained on AffectNet~\cite{mollahosseini2019affectnet}. Thus, the visual backbone is not a face-recognition network trained solely for identity discrimination. The resulting sequence is passed to a Transformer temporal encoder and pooled into
\begin{equation}
h_t^{\mathrm{react}}=\mathrm{Pool}\!\left(E_{\mathrm{temp}}(f_1,\ldots,f_{N_t})\right).
\label{eq:reaction}
\end{equation}
Our primary model uses this temporal representation directly.

To test whether explicitly exposing coarse early-to-late change is necessary, we additionally evaluate an explicit-difference variant. Let $\mathcal E$ and $\mathcal L$ denote the earlier and later valid-frame groups within the same one-second window. We compute
\begin{align}
h^{\mathrm{early}} &= \frac{1}{|\mathcal E|}\sum_{n\in\mathcal E} f_n, &
h^{\mathrm{late}} &= \frac{1}{|\mathcal L|}\sum_{n\in\mathcal L} f_n,\\
\Delta h &= h^{\mathrm{late}}-h^{\mathrm{early}},
\end{align}
and concatenate $\Delta h$ to the temporal representation. This feature is treated as an ablation rather than as the core novelty of ReACT-TTS; Sec.~\ref{sec:ablation} shows that it provides no additional benefit over temporal encoding alone.

\subsection{Target-Conditioned Response Planning}

Visual evidence should not affect every target response equally. ReACT-TTS therefore computes a target-conditioned gate
\begin{equation}
g_t=\sigma\!\left(W_g[h_t^{\mathrm{text}};h_t^{\mathrm{react}}]+b_g\right)
\end{equation}
and fuses linguistic and visual evidence through
\begin{equation}
h_t^{\mathrm{fused}}=h_t^{\mathrm{text}}+g_t\odot W_r h_t^{\mathrm{react}}.
\label{eq:fusion}
\end{equation}
The residual text path preserves the linguistic interpretation when visual evidence is weak, while the gate permits sample-dependent visual modulation. Because $h_t^{\mathrm{text}}$ jointly encodes the dialogue history and target response, the gate determines how much listener evidence is useful for the specific upcoming utterance. From the fused representation, Stage A predicts the target-response emotion, continuous affect, and prosodic attributes,
\begin{align}
\hat{\mathbf y}_t &= \mathrm{softmax}(W_e h_t^{\mathrm{fused}}+b_e),\\
\hat{\mathbf v}_t &= W_v h_t^{\mathrm{fused}}+b_v,\\
\hat{\mathbf p}_t &= W_p h_t^{\mathrm{fused}}+b_p,
\end{align}
where $\hat{\mathbf y}_t$ is the target-emotion distribution, $\hat{\mathbf v}_t\in\mathbb{R}^3$ denotes valence--arousal--dominance (VAD), and
\begin{equation}
\hat{\mathbf p}_t=
[\widehat{\mu}_{\log F0},\widehat{\sigma}_{\log F0},\widehat{E}_{\log},\widehat{R}_{\mathrm{phn}}]
\end{equation}
contains utterance-level mean log-F0, log-F0 standard deviation, mean log-energy, and phoneme rate. The prosodic targets are speaker-normalized as in the response-planning setup. The planner additionally exposes the 256-dimensional response-style embedding $z_t^{\mathrm{style}}$ defined in Eq.~(3), which is the representation passed to Stage C. In all listener-removal baselines, the textual path and prediction heads are kept unchanged; only the listener visual input is removed.

\subsection{Speech Realization}

The speech realization stage uses the same acoustic backbone for the Temporal and Text-only systems. The frozen Stage-A planner produces a 256-dimensional style embedding, which is mapped by a trainable ResponseStyleAdapter into the 256-dimensional acoustic-style space. Separately, a training-speaker reference waveform is encoded with Resemblyzer~\cite{wan2018ge2e} to obtain a 256-dimensional speaker embedding. The style and speaker vectors are concatenated to form a 512-dimensional global condition for Grad-TTS~\cite{popov2021gradtts}:
\begin{align}
z_t^{\mathrm{ac}} &= A_{\mathrm{style}}(z_t^{\mathrm{style}}),\\
c_t &= [z_{s_t};z_t^{\mathrm{ac}}],\\
\hat m_t &= G_{\mathrm{Grad\text{-}TTS}}(x_t,c_t),\\
\hat a_t &= V_{\mathrm{HiFi\text{-}GAN}}(\hat m_t).
\end{align}
The acoustic configuration uses 16-kHz audio, a 1024-point FFT, hop size 160, window size 1024, and 128 mel bins with $f_{\min}=0$ and $f_{\max}=8$~kHz. Monotonic alignment search is used during acoustic training, and a 16-kHz HiFi-GAN vocoder~\cite{kong2020hifigan} converts the generated mel-spectrogram into waveform speech.

\subsection{Training Strategy}

Training is separated into three stages. Stage A trains the response planner. Stage B trains the Grad-TTS acoustic model on the larger MELD acoustic training set using ground-truth style supervision. Stage C freezes Stage A and trains the ResponseStyleAdapter jointly with the acoustic objective so that predicted response style is mapped into the Stage-B acoustic-style space.

For Stage C, the adapter output $z_t^{\mathrm{ac}}$ is additionally aligned to the 256-dimensional Stage-B ground-truth emotion embedding $z_t^{\mathrm{GT}}$ using an equal mixture of mean-squared error and cosine distance,
\begin{equation}
\mathcal{L}_{\mathrm{align}}=0.5\,\mathcal{L}_{\mathrm{MSE}}(z_t^{\mathrm{ac}},z_t^{\mathrm{GT}})
+0.5\,\mathcal{L}_{\mathrm{cos}}(z_t^{\mathrm{ac}},z_t^{\mathrm{GT}}),
\end{equation}
where $\mathcal{L}_{\mathrm{cos}}=1-\cos(\cdot,\cdot)$ denotes cosine distance. The complete Stage-C objective is
\begin{equation}
\mathcal{L}_{C}=\mathcal{L}_{\mathrm{Grad\text{-}TTS}}+\lambda_{\mathrm{align}}\mathcal{L}_{\mathrm{align}},\qquad \lambda_{\mathrm{align}}=0.5.
\label{eq:stagec_loss}
\end{equation}
Ground-truth emotion information is used only as a training target for this alignment and is not available at inference time. The Text-only baseline uses the identical Stage-B/Stage-C acoustic pipeline and differs only by removing listener visual input from Stage A.

\section{Experiments}

\subsection{Dataset and Strict Listener Protocol}

Experiments use MELD~\cite{poria2019meld}, which provides aligned text, audio, video, speaker identity, emotion, and sentiment annotations for multi-party conversations. MELD does not annotate an explicit addressee for every turn, so listener assignment is an important source of uncertainty. We therefore use a precision-oriented protocol and apply the same criteria to train, development, and test data.

A response-planning sample is retained only when the local dialogue contains exactly two speakers, three preceding turns are available, and the target utterance lasts at least one second. The listener reaction is restricted to the one-second interval immediately before the target response. Sixteen frames are requested from this interval. We require listener visibility of at least 0.30, reaction validity of at least 0.50 (at least 8 of 16 requested frames), and face-track identity similarity of at least 0.65. The visually active target speaker is determined using mouth motion, and the remaining tracked dialogue participant is treated as the listener. All thresholds were fixed using the training/development protocol and were not tuned on test performance.

This procedure yields 1,117 training, 116 development, and 261 test samples for response planning (Table~\ref{tab:dataset}), substantially expanding the 102-sample preliminary subset while retaining strict listener filtering. For speech generation, evaluation is further restricted to target speakers for whom a training-split speaker reference is available, leaving 115 development and 252 test utterances. This exclusion is determined by training-reference availability rather than by test-set synthesis quality.

\begin{table}[t]
\centering
\small
\caption{Final MELD protocol. Response-planning samples satisfy the same strict filtering criteria in all splits. Speech-generation evaluation additionally requires a target speaker with a training-split reference.}
\label{tab:dataset}
\begin{tabular}{lrr}
\toprule
Split & Response planning & Speech generation \\
\midrule
Train & 1,117 & -- \\
Dev   & 116   & 115 \\
Test  & 261   & 252 \\
\bottomrule
\end{tabular}
\end{table}

\begin{figure*}[t]
    \centering
    \includegraphics[width=0.96\textwidth]{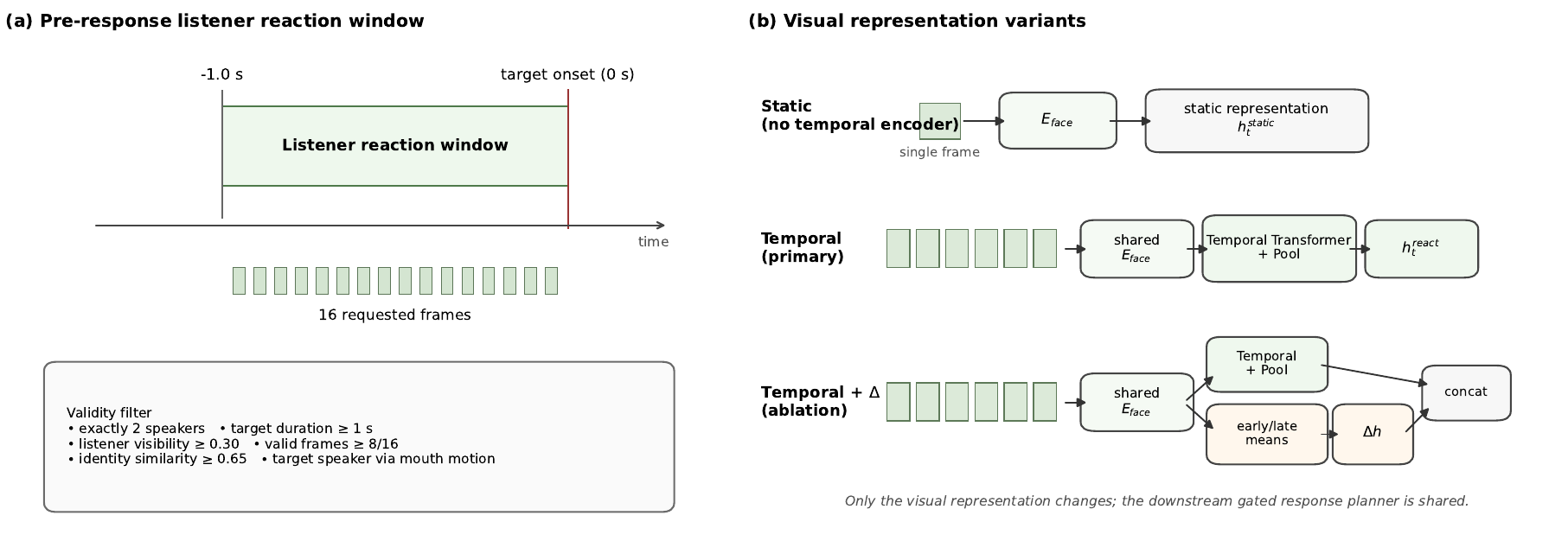}
    \caption{Listener-reaction timing and representation variants. (a) Visual evidence is restricted to the 1-s interval immediately before target onset and retained only after the strict dyadic/track-validity filter. (b) The controlled ablation changes only the visual representation supplied to the same downstream gated planner: a static representation without temporal encoding, the full temporal representation, or the temporal representation augmented with an explicit early-to-late difference.}
    \label{fig:reaction_protocol}
\end{figure*}

\subsection{Implementation and Evaluation Protocol}

The Interaction Context Encoder is initialized from RoBERTa-base~\cite{liu2019roberta}. The facial backbone is a ResNet-18 pretrained on AffectNet~\cite{mollahosseini2019affectnet}, followed by a two-layer Transformer temporal encoder with hidden dimension 256, four attention heads, and dropout 0.1. Stage-A response planning is evaluated using target-emotion accuracy, macro-F1, and VAD concordance correlation coefficient (CCC), averaged across valence, arousal, and dominance. Because class frequencies are highly imbalanced, macro-F1 is treated as the primary classification measure.

For the primary comparison, Text-only and Temporal listener models are evaluated on the fixed 261-sample test set over ten random seeds (42--51). We additionally perform 10,000 bootstrap resamples over test samples to characterize the uncertainty of the paired Temporal-minus-Text difference. The ablation study uses the same five seeds (42--46) for all configurations.

For speech realization, Stage B is trained on 9,988 MELD training utterances (one corrupt media item removed), with seen-speaker development evaluation on 1,071 utterances. The best Stage-B initialization is selected on development loss (seed 42, epoch 23). Stage C freezes Stage A and trains the response-style adapter; the final Temporal and Text-only checkpoints are independently selected by development loss. All reported speech-generation test results use these development-selected checkpoints.

\subsection{Response-Planning Results}
\label{sec:planning}

Table~\ref{tab:style} reports the ten-seed test results. Temporal listener conditioning does not improve raw accuracy (0.4483 vs. 0.4521), but it yields higher mean macro-F1 (0.2582 vs. 0.2489) and CCC (0.2282 vs. 0.2173). The paired macro-F1 difference is $+0.0093\pm0.0184$, with Temporal outperforming Text-only in 7 of 10 seeds.

\begin{table}[t]
\caption{Stage-A response-planning results on the 261-sample test set over ten seeds (42--51), reported as mean $\pm$ standard deviation.}
\label{tab:style}
\centering
\small
\begin{tabular}{lccc}
\toprule
Method & Accuracy $\uparrow$ & Macro-F1 $\uparrow$ & CCC $\uparrow$ \\
\midrule
Text-only & $0.4521\!\pm\!0.0316$ & $0.2489\!\pm\!0.0152$ & $0.2173\!\pm\!0.0463$ \\
Temporal listener & $0.4483\!\pm\!0.0243$ & $\mathbf{0.2582\!\pm\!0.0131}$ & $\mathbf{0.2282\!\pm\!0.0291}$ \\
\bottomrule
\end{tabular}
\end{table}

Bootstrap resampling gives a macro-F1 difference of $+0.0094$ with a 95\% interval of $[-0.0057,+0.0247]$ and $P(\Delta>0)=0.8844$. The corresponding accuracy difference is $-0.0040$ with a 95\% interval of $[-0.0226,+0.0149]$. Because the intervals include zero, we do not interpret these results as statistically significant improvement. Instead, they indicate a modest positive tendency in macro-F1 and VAD concordance, while overall accuracy remains essentially unchanged.

An exploratory class-wise analysis suggests that the contribution of listener information is not uniform across emotions. The largest mean macro-F1 differences occur for surprise ($+0.0391$), sadness ($+0.0225$), and anger ($+0.0158$); Temporal wins 9/10 seeds for surprise and 7/10 for sadness. Neutral is slightly lower and joy is approximately unchanged. Fear and disgust have only 7 and 8 test examples, respectively, and are therefore not interpreted individually. This pattern is consistent with class-dependent complementarity rather than a universal benefit from visual conditioning.

\subsection{Static, Temporal, and Explicit-Difference Ablation}
\label{sec:ablation}

To isolate the contribution of static appearance, temporal modeling, and the explicit difference feature, we conduct a controlled five-seed ablation using the representation variants in Fig.~\ref{fig:reaction_protocol}(b). Table~\ref{tab:ablation} reports the results. A static face representation improves modestly over Text-only on average, but the strongest result is obtained by the Temporal model without an explicit $\Delta$ feature (0.2558 macro-F1). Adding the explicit difference reduces the mean to 0.2407.

\begin{table}[t]
\caption{Five-seed fair ablation (seeds 42--46). The temporal model without an explicit difference feature performs best in Macro-F1.}
\label{tab:ablation}
\centering
\small
\begin{tabular}{lc}
\toprule
Configuration & Macro-F1 $\uparrow$ \\
\midrule
Text-only & $0.2367 \pm 0.0021$ \\
Static face & $0.2429 \pm 0.0272$ \\
Temporal + explicit $\Delta$ & $0.2407 \pm 0.0191$ \\
\textbf{Temporal, no $\Delta$} & $\mathbf{0.2558 \pm 0.0135}$ \\
\bottomrule
\end{tabular}
\end{table}

These results shift the interpretation of the method away from hand-crafted embedding differences. The AffectNet-pretrained backbone provides expression-oriented frame features, while the temporal encoder models their evolution directly. The explicit early-to-late difference does not provide additional benefit in our setting, so the no-$\Delta$ Temporal configuration is used as the primary ReACT-TTS planner.

\subsection{Mismatched-Listener Counterfactual Intervention}
\label{sec:mismatch}

To test whether performance arises from listener-specific information rather than generic visual regularization, we perform a counterfactual input intervention at test time. For each batch, the correct listener sequence is replaced by another sample's listener sequence using within-batch cyclic reassignment; dialogue history, target text, and all non-visual inputs remain unchanged. This deterministic reassignment avoids tuning a favorable mismatch for individual examples.

Table~\ref{tab:mismatch} shows that the correct listener yields 0.2582 macro-F1 compared with 0.2526 for the mismatched condition, a mean difference of approximately $+0.0055$. Across seeds, the correct listener wins/ties/loses in approximately 6/2/2 cases. Test-sample bootstrap gives a 95\% interval of $[-0.0027,+0.0135]$ and $P(\Delta>0)=0.911$.

\begin{table}[t]
\caption{Correct-versus-mismatched listener intervention over ten seeds. Mismatched reactions are produced by within-batch cyclic reassignment while text context and target response remain fixed.}
\label{tab:mismatch}
\centering
\small
\begin{tabular}{lc}
\toprule
Listener input & Macro-F1 $\uparrow$ \\
\midrule
Mismatched listener & $0.2526 \pm 0.0117$ \\
Correct listener & $\mathbf{0.2582 \pm 0.0131}$ \\
\bottomrule
\end{tabular}
\end{table}

The interval again includes zero, so this experiment does not establish a significant or causal advantage. It nevertheless provides complementary evidence that the planner is sensitive to which listener reaction is paired with the conversation: replacing the reaction while keeping the linguistic response fixed tends to reduce macro-F1.

\subsection{End-to-End Speech Realization}
\label{sec:generation}

We next evaluate whether the predicted response representation can be connected to an end-to-end speech generator. The frozen Stage-A style representation is mapped through the ResponseStyleAdapter and combined with a Resemblyzer target-speaker embedding to condition Grad-TTS. Temporal and Text-only systems use the same acoustic backbone and differ only in the Stage-A visual input.

Initial synthesis at the default length scale 1.0 produced speech that was substantially shorter than the reference (mean generated duration about 1.58--1.59~s versus 3.55~s for the reference). We therefore selected the inference length scale on the 115-sample development set, without consulting test results. Among scales 1.5, 1.75, 2.0, and 2.25, scale 1.5 gave the lowest development WER (approximately 1.03) and was fixed for final inference, despite longer scales producing duration ratios closer to one.

Table~\ref{tab:speech} reports objective evaluation on 252 test utterances. Temporal conditioning produces slightly higher speaker similarity (0.6059 vs. 0.5976), while Text-only is slightly better in WER (1.0320 vs. 1.0384) and CER (0.8824 vs. 0.8967). Thus, the generated-speech experiment demonstrates that the response plan can be carried through the acoustic pipeline, but it does not support a claim that listener conditioning improves intelligibility or overall acoustic quality.

\begin{table}[t]
\caption{Objective speech-generation evaluation on 252 test utterances at length scale 1.5, selected on the development split. Lower WER/CER is better; higher speaker similarity is better.}
\label{tab:speech}
\centering
\small
\begin{tabular}{lccc}
\toprule
Condition & WER $\downarrow$ & CER $\downarrow$ & Spk. Sim. $\uparrow$ \\
\midrule
Text-only & $\mathbf{1.0320}$ & $\mathbf{0.8824}$ & $0.5976$ \\
Temporal listener & $1.0384$ & $0.8967$ & $\mathbf{0.6059}$ \\
\bottomrule
\end{tabular}
\end{table}

As a diagnostic, we also synthesized speech from Stage B using ground-truth emotion/style conditioning. Intelligibility remained poor, indicating that the high WER/CER cannot be attributed solely to the Stage-C response-style adapter. We therefore treat the acoustic backbone and duration modeling as important limitations of the current realization stage rather than using synthesis quality as evidence for the response-planning claim.

\paragraph{Perceptual contextual appropriateness.}
We additionally conduct a preference study with 20 speech researchers holding an M.S. degree or higher in AI-related fields. Given the preceding conversation, target text, and pre-response listener reaction, evaluators compared anonymized Speech A/B realizations of the same text and selected which delivery better fit the listener reaction and conversational context, with a no-preference option. Across collected judgments, \textbf{76\%} preferred Temporal, 9\% Text-only, and 15\% reported no preference/equal appropriateness. This evaluates contextual appropriateness rather than general naturalness, which remains confounded by the shared acoustic limitations in Table~\ref{tab:speech}.

\section{Discussion and Limitations}

Across ten seeds, Temporal listener conditioning shows a positive tendency in macro-F1 and CCC, but the paired bootstrap interval includes zero and raw accuracy is essentially unchanged. We therefore interpret pre-response listener behavior as a complementary, class-dependent planning cue rather than a universal improvement. The five-seed ablation further localizes the useful visual signal: an AffectNet-pretrained, expression-oriented backbone with temporal encoding achieves the highest mean macro-F1, whereas the explicit early-to-late $\Delta$ adds no benefit. Thus, temporal listener dynamics---not an engineered difference feature---form the core visual contribution.

Listener assignment remains imperfect because MELD lacks explicit addressee annotations. The two-speaker, mouth-motion, visibility/validity, and identity filters reduce ambiguity but cannot guarantee communicative intent and may favor clearly visible faces. The mismatched-listener intervention provides complementary evidence that the correct sequence is more useful on average, but its confidence interval also includes zero; it is a controlled network-input intervention, not evidence of a causal mechanism in human communication. Explicit dyadic addressee annotations and richer action-unit or expression trajectories are important future tests.

Speech realization remains the main practical limitation. Both systems have high WER/CER, and the weak Stage-B oracle diagnostic points to the acoustic backbone and duration model rather than listener conditioning alone. Accordingly, generation is used to establish feasibility, not superior synthesis quality. The researcher preference study nevertheless shows a clear descriptive preference for Temporal contextual appropriateness; it does not evaluate naturalness, which requires a stronger synthesis backbone and dedicated study.

Direct numerical comparison with prior conversational or visually conditioned TTS remains difficult because datasets and conditioning definitions differ. Evaluation on shared conversational benchmarks remains important future work.

\section{Conclusion}

We presented ReACT-TTS, which uses the listener's one-second pre-response facial behavior as complementary context for explicit response planning. On a strict dyadic MELD protocol, Temporal conditioning shows a modest positive tendency in macro-F1 and CCC and achieves the highest mean macro-F1 among the controlled static and explicit-$\Delta$ variants. Correct-versus-mismatched input results further suggest listener-specific information use, while contextual-appropriateness judgments favor Temporal delivery. The planned representation can condition Grad-TTS end to end, although current synthesis quality remains limited. These results motivate pre-response listener dynamics as a useful planning signal while separating that question from acoustic-generation quality.

\bibliographystyle{splncs04}
\bibliography{references}

\end{document}